\documentclass[a4paper, 10pt, conference]{IEEEtran}
\IEEEoverridecommandlockouts
\usepackage{cite}
\usepackage{amsmath,amssymb,amsfonts}
\usepackage{algorithmic}
\usepackage{graphicx}
\usepackage{textcomp}
\usepackage{xcolor}
\usepackage{hyperref}
    
\begin{document}

\title{Uncovering variability in human driving behavior through automatic extraction of similar traffic scenes from large naturalistic datasets
\thanks{This work was funded by Nissan Motor Co. Ltd.}}

\date{February 2022}

\author{\IEEEauthorblockN{Olger Siebinga}
\IEEEauthorblockA{\textit{Human-Robot Interaction, CoR, 3ME} \\
\textit{Delft University of Technology}\\
Delft, the Netherlands \\
0000-0002-5614-1262}
\and
\IEEEauthorblockN{Arkady Zgonnikov}
\IEEEauthorblockA{\textit{Human-Robot Interaction, CoR, 3ME} \\
\textit{Delft University of Technology}\\
Delft, the Netherlands \\
0000-0002-6593-6948}
\and
\IEEEauthorblockN{David Abbink}
\IEEEauthorblockA{\textit{Human-Robot Interaction, CoR, 3ME} \\
\textit{Delft University of Technology}\\
Delft, the Netherlands \\
0000-0001-7778-0090}
}

\maketitle
\begin{abstract}
    Recently, multiple naturalistic traffic datasets of human-driven trajectories have been published (e.g., highD, NGSim, and pNEUMA). These datasets have been used in studies that investigate variability in human driving behavior, for example for scenario-based validation of autonomous vehicle (AV) behavior, modeling driver behavior, or validating driver models. Thus far, these studies focused on the variability on an operational level (e.g., velocity profiles during a lane change), not on a tactical level (i.e., to change lanes or not). Investigating the variability on both levels is necessary to develop driver models and AVs that include multiple tactical behaviors. To expose multi-level variability, the human responses to the same traffic scene could be investigated. However, no method exists to automatically extract similar scenes from datasets. Here, we present a four-step extraction method that uses the Hausdorff distance, a mathematical distance metric for sets. We performed a case study on the highD dataset that showed that the method is practically applicable. The human responses to the selected scenes exposed the variability on both the tactical and operational levels. With this new method, the variability in operational \textit{and} tactical human behavior can be investigated, without the need for costly and time-consuming driving-simulator experiments.
\end{abstract}

\section{Introduction}
In recent years, multiple open-access naturalistic datasets have been published. Some of these datasets are constructed by first recording videos of traffic with mounted cameras (e.g., NGSIM~\cite{NGSIM2016}) or drones (e.g., highD~\cite{Krajewski2018} and pNEUMA~\cite{Barmpounakis2020}). Image recognition techniques are then used to extract trajectory data from these videos. Such datasets contain trajectories for all vehicles that pass through a specific area. 

Researchers have used these datasets for multiple purposes, among which: scenario-based validation of autonomous vehicle (AV) behavior (see~\cite{Riedmaier2020} for a review), modeling and predicting driver behavior (e.g.,\cite{Mahajan2020,Krajewski2018a,Schwarting2019}), and validating driver models to be used in autonomous vehicles (e.g.,\cite{Siebinga2022b}). In all these applications, the \textit{variability} (the range of human behaviors that can be expected in response to a given situation, sometimes referred to as uncertainty) in driver behavior is relevant.

Currently, variability is mostly regarded on the level of \textit{operational} driving behavior (e.g., \cite{Ossen2011, Kurtc2020, Thiemann2008, Mahajan2020, Krajewski2018a}). \textit{Operational} driving behavior considers the execution of a maneuver~\cite{Michon1985}, for example a lane change. However, variability does also exist on the \textit{tactical} level, that is, in the choice of maneuver when a driver responds to a traffic scene~\cite{Michon1985}. For instance, some drivers might respond to a slower-moving vehicle in their lane by overtaking it, while others will brake in the same situation.

Understanding variability on both the operational and the tactical level is important for assessing the human-likeness and acceptability of AV behavior, and also for validating human driver models used in AVs~\cite{Siebinga2022b}. The reason is that both these applications must consider all possible tactical behaviors under given conditions. Traditional driver models on the other hand, mostly target a specific tactical behavior (e.g., car following in the Intelligent Driver Model~\cite{Treiber2000}), thus for their application, only operational variability is relevant. When designing driver models that describe multiple tactical behaviors, the variability in tactical behavior also needs to be understood. 

To study driver behavior variability to its full extent, similar traffic scenes have to be (automatically) extracted from the previously mentioned datasets in order to compare the human responses to these scenes. However, most automatic extraction methods select traffic \textit{scenarios} (see~\cite{Riedmaier2020} for a review) not traffic \textit{scenes}. According to Ulbrich et al.~\cite{Ulbrich2015}, a scenario "describes the temporal development between several scenes in a sequence of scenes..", where "a [traffic] scene describes a snapshot of the environment including the scenery and dynamic elements.." (dynamic elements in the discussed datasets are (human-driven) vehicles). These definitions show that (extracted) traffic scenarios include part of a trajectory. Trajectories that are similar describe the same tactical behavior in most cases. Thus, selecting similar traffic scenarios implicitly means selecting similar tactical responses. Some other approaches even explicitly extract data corresponding to a pre-specified tactical behavior (e.g. lane changes in~\cite{Krajewski2018a, Siebinga2022b}).

\begin{figure*}[ht]
    \centering
    \includegraphics[width=\textwidth]{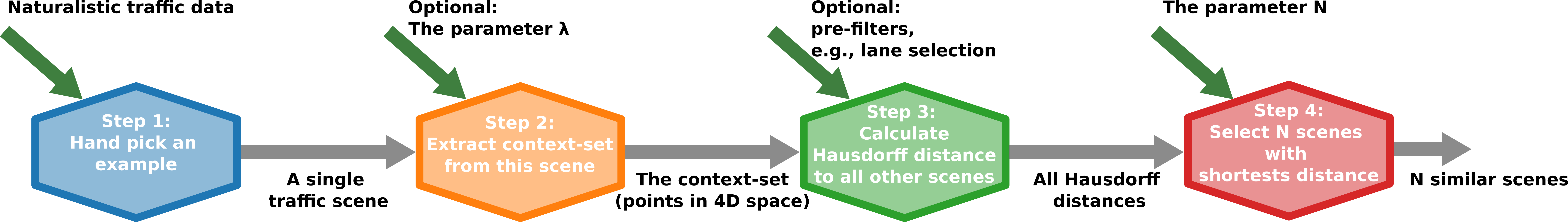}
    \caption{The four steps of the proposed method to find similar scenes in a naturalistic traffic dataset. In the first step, the scene of interest (of which similar instances should be found) is defined by manually selecting an example of this scene from the naturalistic dataset. The second step consists of extracting the traffic context from this example and converting it to a mathematical set of points --- the context set. We use the scaling parameter $\lambda$ in this conversion because we target highway traffic, but this is optional. The distance between this set and all other sets present in the data is calculated in step three using the Hausdorff distance. Optional pre-filtering steps can be used here to reduce the computational load of the method. Finally, in step four the $N$ scenes with the shortest distance to the example can be selected.}
    \label{fig:steps}
\end{figure*}

These existing approaches expose the variability in the operational execution of a given tactical maneuver but disregard the variability in tactical behavior. Furthermore, including trajectories as part of the automatically extracted data conflates the initial traffic scene (i.e., what a driver is responding to) and the driver's response itself. This makes it more difficult to investigate the full extent of variability in human responses to a specific initial traffic scene.

A method to automatically extract similar traffic scenes from naturalistic datasets would support studies into driver behavior variability on both operational and tactical levels. With such a method, the trajectories in response to an initial scene can be studied, both in terms of operational and tactical characteristics. However, to the best of our knowledge, all the methods that have been proposed to automatically extract traffic \textit{scenarios} from the data cannot extract traffic \textit{scenes}.

This paper proposes a method to automatically extract similar traffic scenes from large naturalistic datasets (for a schematic overview, see Figure~\ref{fig:steps}). We introduce the concept of \textit{traffic context} to specify the part of the initial traffic scene that is relevant for comparing human responses. We define \textit{traffic context} as all positions and velocities of all surrounding vehicles at a given time. Compared to the complete traffic scene, the traffic context excludes scenery and the state of the ego vehicle. Therefore, the traffic context represents the aspects of the scene the ego-vehicle is responding to.

We use a distance metric to express the difference between traffic contexts. With such a measurable distance, it becomes possible to automatically find scenes that are similar to a manually-selected example. However, such a distance metric for traffic scenes does not readily exist. Therefore, we propose to convert the traffic context to a mathematical set and use the Hausdorff distance for mathematical sets~\cite{Hausdorff1914}. An implementation of our method is provided on GitHub\footnote{\href{https://github.com/tud-hri/hausdorffsceneextraction}{github.com/tud-hri/hausdorffsceneextraction}} as an extension of the traffic visualization software TraViA~\cite{Siebinga2021}. We validated the method in a case study using the highD dataset, where we show that this method is practically applicable and provides insight into the operational and tactical variability of driver behavior. 

\section{Proposed Method}
Our proposed method consists of four steps (Figure~\ref{fig:steps}). These steps are briefly introduced in this section and explained in more detail when applied in the case study. In the first step, one should manually select an example of the scene of interest from the dataset. This example represents the traffic scene of interest of which multiple instances should be found. The traffic context from this example is converted to a mathematical set (the \textit{context set}) in the second step. After that, the Hausdorff distance is used to determine the distance between the traffic context in the selected scene and all other scenes in the dataset. Finally, one selects the $N$ contexts with the shortest distance to the example. The resulting scenes are the scenes with traffic contexts most similar to the example across the whole dataset.

\begin{figure}[h]
    \centering
    \includegraphics[width=0.6\linewidth]{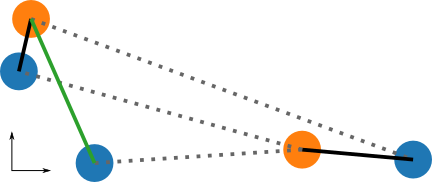}
    \caption{A visual example of how to calculate the Hausdorff distance from the blue set $A$ to the orange set $B$ in a 2-dimensional space. First, for every point in the blue set, find the closest (Euclidean) distance to any point in the orange set. These are shown here as solid lines. Then, select the longest of these minimum (solid-line) distances, this is the Hausdorff distance. This distance is here shown as the green solid line.}
    \label{fig:hausdorff_example}
\end{figure}

The Hausdorff distance is a distance metric for mathematical sets proposed by Felix Hausdorff in 1914~\cite{Hausdorff1914}. It can be used to express the distance between two non-empty compact sets. For two sets $A$ and $B$, the Hausdorff distance represents the maximum distance between any point in set $A$ and the closest point in set $B$ (Figure~\ref{fig:hausdorff_example}). It can be used to describe the similarity between two sets, even if these sets have a different number of points. This practically means that it can be used to compare scenes with a different number of vehicles. 

More extensive (and formal) explanations of the Hausdorff distance and how it can be calculated can be found in the literature (e.g.,~\cite{Rucklidge1996}) and online (e.g.,~\cite{WikipediaHausdorff}). Mathematically, the (directed) Hausdorff distance between two sets $A$ and $B$ is defined as:

\begin{equation}
    h(A, B) = \max_{a \in A} \{ \min_{b \in B} \{ d(a, b)\}\}
\end{equation}

Where we use the Euclidean distance between points $a$ and $b$ for distance $d(a, b)$. The directed Hausdorff distance from set $A$ to $B$ is not equal to the distance from $B$ to $A$ (i.e., $h(A, B) \neq h(B, A)$). Therefore we use the general Hausdorff distance H:

\begin{equation}
    H(A, B) = \max\{ h(A, B), h(B, A)\}
\end{equation}

\section{Case Study: Methods}
\label{sec:methods}

In this case study, we make use of the highD dataset~\cite{Krajewski2018} to show the potential of our proposed method. The highD dataset consists of traffic data recorded in Germany at 6 different highway locations. The dataset is made up of 60 independent recordings. To visualize the data and generate the images used in this work, we used the TraViA visualization software~\cite{Siebinga2021}. The source code implementing the proposed method is publicly available as an extension to TraViA~\cite{Siebinga2022}.

\begin{figure*}[ht]
    \centering
    \includegraphics[width=\textwidth]{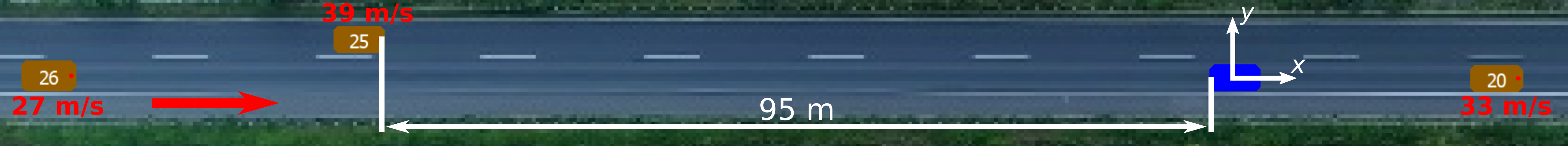}
    \caption{The hand-picked example of the traffic scene of interest, as used in step 1 of our case study (highD dataset 1, frame 379, ego vehicle id 21). The blue vehicle is the ego vehicle, with the red arrow indicating the driving direction.  The orange vehicles (id 25,26,20) denote other traffic participants, making up the traffic context, all driving in the same direction on the two-lane highway. In this example, the driver could decide to stay behind the vehicle it is currently following (id 20), but could also decide to accelerate and overtake that vehicle. The white arrows depicting x and y show the ego vehicle's reference frame. The gap between vehicle 25 and the ego vehicle is 95 meters. The numbers in red denote the longitudinal velocities of the surrounding vehicles.}
    \label{fig:example_situation}
\end{figure*}

\subsection{Step 1: select an example}
The first step of the proposed method is to select an example of the scene, to which the variability in responses is the topic of research. We will refer to this scene as \textit{the traffic scene of interest}. This example should be at a specific point in time, seen from the perspective of a selected ego vehicle. For the highD dataset, this means that the example can be fully defined by a combination of three numbers: a dataset id, a vehicle id, and a frame number. For this case study, we have selected the example as depicted in Figure~\ref{fig:example_situation}. This scene can be found in dataset 1, frame 379 with ego vehicle id 21. 

This example was selected because the driver of the ego vehicle (blue, id 21) can respond to this scene in multiple (tactical) ways as illustrated in Figure~\ref{fig:example_situation}: the driver could decide to stay behind the vehicle it is currently following (id 20), but could also decide to accelerate and overtake that vehicle. The headway between the following vehicles (ids 25 and 26) and the ego vehicle is large enough ($95~m$ and $128.7~m$) to allow the ego vehicle to change lanes, but small enough to expect some effect of their presence on the ego vehicle's behavior.

\subsection{Step 2: extract the context set}
The second step of our proposed method is to convert the traffic context to a mathematical set of 4-dimensional points. We will refer to this set as the context set consisting of context points. There is one context point for every surrounding vehicle. The context set can contain any number of context points, depending on the number of surrounding vehicles that are assumed to be part of the traffic context. In our case study, we used the definitions provided by highD to determine the vehicles that make up the traffic context. In the highD dataset, 8 positions for surrounding vehicles are reported for every ego vehicle. We assume these surrounding vehicles make up the traffic context.

To convert the state of the surrounding vehicles to context points, the 2-dimensional position of each vehicle is expressed in the ego vehicle's reference frame. The 2-dimensional absolute velocities (expressed along the same axes) of each vehicle are then concatenated to the relative positions. The result is a 4-dimensional point per surrounding vehicle. In mathematical form, a single context point representing a single surrounding vehicle can be expressed as

\begin{equation}
\label{eq:context_point}
p_i = [x_i, y_i, v_{x, i}, v_{y, i}],
\end{equation}
where $i$ denotes the $i^{th}$ surrounding vehicle, $x$ and $y$ denote the center $x$-position and $y$-position of the vehicle relative to the ego vehicle and $v_x$ and $v_y$ the velocities in $x$ and $y$ direction.

One potential problem with defining the context points as is done in Equation~\ref{eq:context_point} is that it regards differences in longitudinal and lateral position equally. However, on highways such as in the highD dataset, a small (e.g. 4 m) lateral difference in position can mean a vehicle is driving in another lane. This would substantially change the scene. While the same difference in longitudinal position would have a much smaller effect. To account for this difference, we introduce the parameter $\lambda$ to scale the lateral dimension of the context points. This parameter should be estimated to reflect the relative importance of longitudinal and lateral position and velocity differences. With the new parameter $\lambda$, the definition of the context points becomes

\begin{equation}
\label{eq:context_point_lambda}
p_i = [x_i, \lambda y_i, v_{x, i}, \lambda v_{y, i}].
\end{equation}

In our case study, we have assumed $\lambda=10.0$. This corresponds to the notion that a $1$ meter change in lateral position of a surrounding vehicle is equally important as a $10$ meter change in longitudinal position. 

\subsection{Step 3: apply the Hausdorff distance}
Now that the traffic context has been represented as a mathematical set, we can use the Hausdorff distance to compare different context sets. This step of the proposed method requires the Hausdorff distance to be calculated between the context set of the selected example and the context sets for all possible combinations of frame number and vehicle id in the dataset. For the highD dataset, there are $39.7 \times 10^6$ such combinations. Because this is a very large number of distances to be calculated, we will reduce it by filtering the relevant vehicles before calculating the distances.

When searching for scenes with similar traffic contexts, an important aspect is the lane the ego vehicle is driving in. This determines where surrounding vehicles can be present and in which directions the ego vehicle can change lanes (e.g., a vehicle driving in the centre lane can go both left and right, but a vehicle driving in the left lane can only change lanes to the right). For that reason we only consider vehicles driving in the same lane as the ego vehicle in the selected example. We consider 4 possible lanes: the left lane, the centre lane, the right lane, and the merging lane. We determine the lane in the selected example (e.g., for Figure~\ref{fig:example_situation}: the right lane) and only use the vehicle frame combinations from the dataset where the vehicle drives in the same lane. For our case study, this leaves $12,515,286$ vehicle-frame combinations. 

If the resulting number of distances to be calculated is still too large after applying this filter, one could consider down-sampling the frames. Depending on the specific frame rate of a chosen dataset, one could assume that the traffic context does not substantially differ within a certain number of frames and therefore only look at a subset of all frames. This would reduce the number of distances to be calculated even further. In our case study, this was not necessary because the resulting number of required distance calculations proved to be feasible. The calculation time on a high-end desktop (9\textsuperscript{th} Gen Intel i9 8-core) was 3 hours, on a consumer-grade laptop (8\textsuperscript{th} Gen Intel i7 4-core) it took 5 hours.

\subsection{Step 4: obtain scenes}
When all Hausdorff distances are calculated, the scenes in the dataset that are closest to the example can easily be obtained by selecting the $N$ shortest distances. The only caveat here is that consecutive data frames are very similar, which results in groups with the same vehicle id and many consecutive frame numbers having very similar (short) Hausdorff distances to the example of the scene of interest. This problem can be accounted for by sorting all results based on the shortest distance only keeping the highest entry for every vehicle. Selecting the top $N$ entries from the resulting table yields the final result.

\begin{figure*}[h!]
    \centering
    \includegraphics[width=\textwidth]{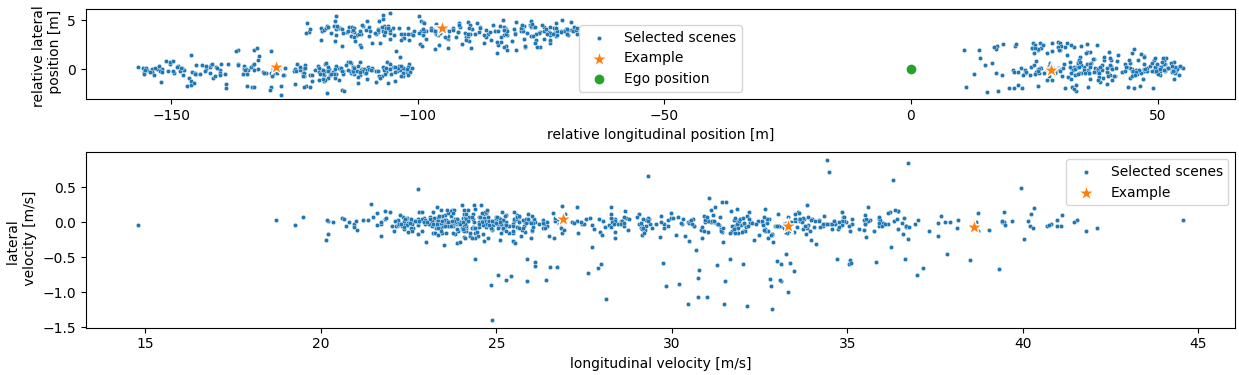}
    \caption{The spread of the context points representing the results of the case study obtained after the final step. The top plot shows the positions of surrounding vehicles relative to the ego vehicle (see Figure~\ref{fig:example_situation} for the frame definition) where the starts represent the scene of interest. The ego vehicle always drives in the right-most lane. The bottom plot shows the absolute velocities of the surrounding vehicles (denoted by the stars). The ego vehicle's velocity is not regarded as part of the traffic context, so it differs for all scenes and is not depicted here. The stars represent the context set extracted from the selected example (Figure~\ref{fig:example_situation}). The blue dots represent the $250$ closest context sets that were automatically extracted from the highD dataset.}
    \label{fig:situation_spread}

    \centering
    \includegraphics[width=0.95\textwidth]{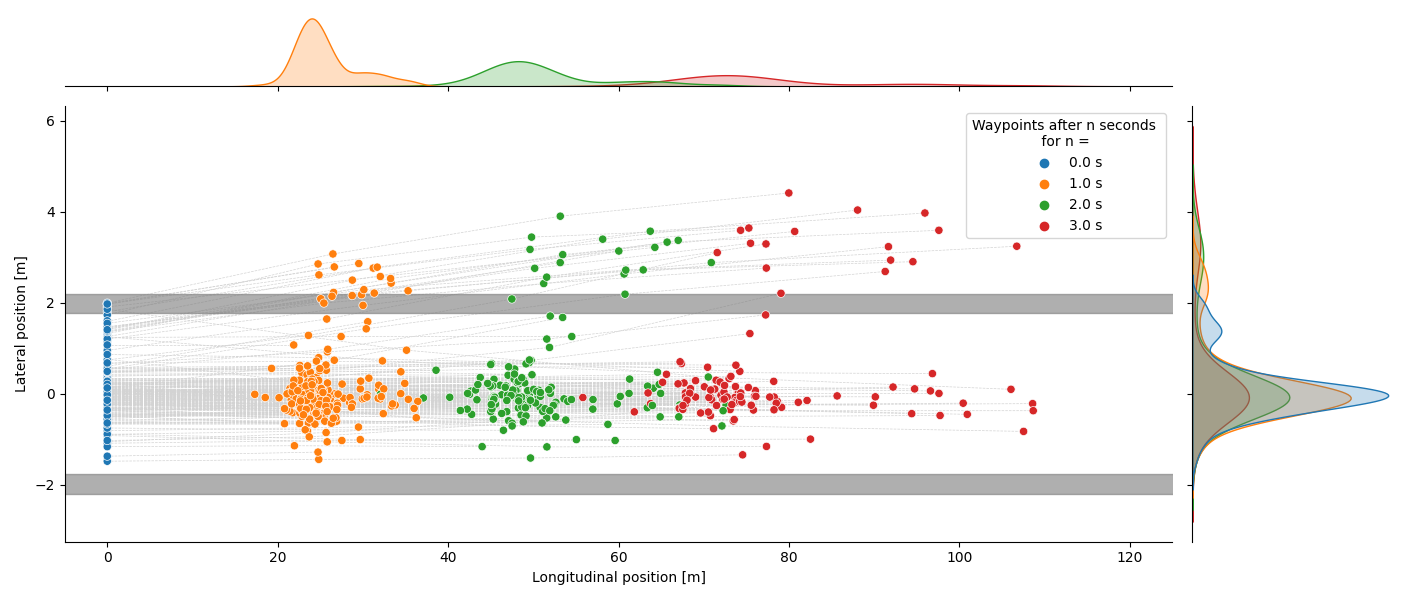}
    \caption{The variability in driver responses (driven trajectories) as they evolve from the 250 traffic scenes with similar traffic context automatically extracted from the highD dataset (represented in Figure~\ref{fig:situation_spread}). The lateral positions are normalized such that $0~m$ indicates the center of the original (right-most) driving lane. Lane widths differ slightly within the highD dataset but are approximately $4~m$. The horizontal grey bars show the range in which all lane markings fall. Blue dots depict the drivers' initial positions and grey lines depict their individual trajectories - with markers for 1 (orange), 2 (green), and 3 (red) seconds. Distributions on longitudinal and lateral vehicle behavior are estimated and shown on the top and right sides of the figure. The figure illustrates tactical variability: some vehicles (n=19) make a lane change (the red dots with substantial positive lateral positions) while others keep car following in the original lane. Operational variability can also be observed in position and velocity for both the lane-changers and the car-followers.}
    \label{fig:hausdorff_results}
\end{figure*}

\section{Case Study: Results}
We used the proposed method in a case study to extract 250 scenes with a similar traffic context from the highD dataset. The hand-picked example of the scene of interest in step 1 is illustrated in Figure~\ref{fig:example_situation}. The proposed method resulted in 250 scenes, where the traffic context is closest to this example. The spread of the resulting context points is shown in Figure~\ref{fig:situation_spread}. Of the 250 found scenes depicted in Figure~\ref{fig:situation_spread}, 233 contain precisely 3 surrounding vehicles, the same number as in the scene depicted in Figure~\ref{fig:example_situation}. The other 17 scenes contain 4 surrounding vehicles.

Figure~\ref{fig:situation_spread} shows that the proposed method for automatically selecting scenes from a dataset succeeds in selecting context sets that are similar to the traffic context of the scene of interest. Note that the three clusters in this figure are not three independent distributions. The Hausdorff distance between sets can be interpreted as a trade-off between the points in a set. If one point is far away from the example, the other two need to be closer to result in a short Hausdorff distance. Therefore, the points within one set cannot be seen as samples from independent distributions. 

Figure~\ref{fig:situation_spread} also shows that the resulting spread is larger in the longitudinal direction than in the lateral direction. For example, in longitudinal positions, the maximum difference between the found sets and the selected example is approximately $25~m$ where the maximum lateral deviation is approximately $2~m$. These values correspond to the used $\lambda$ value of $10$. 

The variability in the results does depend on the amount of data and the scene of interest. The proposed method finds the closest available sets, so if the example represents a more common scene or the dataset to search is larger, lower variability in the found context sets can be expected. The variability in results can also be reduced by selecting fewer context sets i.e. select the $N=100$ closest set instead of the $N=250$, but this is a trade-off with the power of the resulting variability estimation. 

Among other use cases, these results can be used for research targeting the variability in human responses to similar traffic contexts. To illustrate the utility of the results, Figure~\ref{fig:hausdorff_results} shows these human responses. 26 drivers (10\%) responded to this scenario by changing lanes within 3 seconds, 62 drivers (25 \%) slowed down by more than 1 \% of their initial velocity, and the other 162 drivers (65 \%) did not slow down or change lanes within the 3 second period.

The figure also shows kernel density estimations of the longitudinal and lateral distributions for multiple points in time. These estimated distributions could be used to validate driver models that make predictions in the form of distributions. The figure illustrates two potential benefits of the proposed method: the method can be used to extract scenes to which humans respond with different tactical behaviors, and  distributions of human behavior can be estimated from the responses to these selected scenes. 

\section{Discussion}

In this paper, we propose a novel method to automatically extract similar traffic scenes from large naturalistic datasets. In a case study on the highD dataset, we showed that the proposed method is practically applicable and provides insightful results that expose the operational \textit{and} tactical variability in human responses to similar traffic scenes. Therefore, our proposed method can be a valuable tool for the development of autonomous vehicles and traffic systems that incorporate human responses in their control decisions. Also, the case study showed that humans respond to similar traffic scenes with different tactical behaviors (some change lanes while others stay in their initial lane).

One approach closely related to our method is that of clustering scenarios. As discussed in the introduction, obtaining similar scenarios serves a different use case than extracting similar scenes. However, clustering requires a distance metric which makes it comparable to our method. There are two specific trajectory clustering methods that bear resemblance to our approach. In~\cite{Atev2010}, the same distance metric is used as in our approach: the Hausdorff distance. However, in their approach, it is used to determine the distance between two trajectories by regarding the waypoints as a set while we convert the traffic context to a mathematical set. In~\cite{Kerber2020}, another distance metric for scenes is proposed based on a grid around the ego vehicle and the longitudinal distances to other vehicles. Although similar scenes can indeed be found using only longitudinal distance, our method based on the Hausdorff distance is more complete because it also takes into account the lateral positions and longitudinal and lateral velocities of the surrounding vehicles.

Using naturalistic traffic datasets is not the only way to investigate variability in human responses to the same scene, driving-simulator experiments are a well-established alternative. In a driving simulator, multiple participants can be subjected to exactly the same scene with the same traffic context. However, naturalistic data should be used for some applications, for instance when validating human driver models for autonomous vehicles~\cite{Siebinga2022b}. In other cases, a large diversity of drivers might be needed (e.g., when interested in behavior across the population). This would make driving-simulator experiments time-consuming and expensive. For those reasons, our proposed method based on naturalistic data to study human responses to similar traffic contexts is a valuable new approach.

The proposed approach has four main limitations, some of which can be addressed in future work. First, there is no measure to determine how similar two traffic scenes are from a human perspective. This means that the magnitude of similarities and differences between the selected scenes, and thus the method's performance, cannot be quantified. The best way to construct such a measure would be to collect similarity ratings from humans by letting them experience selected pairs of scenes from the dataset. 

Second, the dimensions of vehicles are not taken into account for the traffic context. This could be addressed in a post-processing step if these dimensions are deemed important to answer one's research question. This might, however, limit the amount of extracted scenes. Third, the initial velocity of the ego vehicle is ignored. This was done purposefully because we argue that the initial velocity of the ego vehicle is part of the human response, not of the traffic context. This is a limitation when the resulting data is used to validate driver models that do take this information into account. Including the ego vehicle's velocity could be done by adding the ego vehicle as an extra context point to the context set at position $(0.0, 0.0)$. 

Finally, we presented no systematic approach to determine the parameter $\lambda$. The main reason is that the relative importance of the longitudinal and lateral positions of other vehicles can depend on a number of factors, such as the environment, vehicles' velocities, road dimensions, and the targeted scene. We would like to point out that we needed the $\lambda$ parameters because we used highway data. In other environments with a single lane per direction, such as in inner-city traffic, the $\lambda$ parameter would not be needed. 

To investigate the sensitivity of the $\lambda$ parameter we repeated our selection procedure with the $\lambda$-value increased and decreased by 10\% ($\lambda=9.0$ and $\lambda=11.0$). We evaluated the results by comparing the found dataset and vehicle IDs between the original and new $\lambda$ values. The results for the lowered $\lambda$ deviated for 15 vehicles (6\%) and for the increased $\lambda$ for 12 vehicles (5\%). We thus conclude that our proposed method is not extremely sensitive to the choice of $\lambda$, and that it can be safely estimated manually.

Besides the limitations, there are some possibilities for extending the proposed method. It could be extended to include multiple types of traffic users (e.g., vehicles, pedestrians, and cyclists). To do this, every group of traffic users should be converted to an individual set. The distances of all sets can then be summed in step 3 to find the closest scenes. Besides that, contextual information could be regarded in the post-processing step. This would allow for the inclusion of factors such as weather or lighting conditions (if this information is available with the data).

Furthermore, we used the highD dataset in our case study, but the method itself is suitable for use with all trajectory data (e.g., with the pNeuma or NGSim datasets). The main advantage of the highD dataset is that includes information about the surrounding vehicles. This is a pre-processing step that has to be performed on other datasets before they can be used with the proposed method.

In this paper, we have shown a case study on a single example from the highD dataset. Although we believe it is an illustrative example, it only shows the results of our method for a single scene. To verify if the method is generalizable, we repeated the procedure for other example scenes. However, due to the page limit, we did not share those results here. To aid in the reproduction of these results and to enable the replication (including the generation of figures) on other scenes from the highD dataset, we openly share the source code of our method~\cite{Siebinga2022}. Future studies can also use this code to systematically investigate the use of our method for different applications and other traffic datasets.

\section{Conclusion}
We conclude that:

\begin{itemize}
    \item When a set of comparable traffic scenes needs to be extracted from a large naturalistic dataset (e.g. for human factors analyses), our proposed methodology offers an automated and repeatable approach. We demonstrate our method on a HighD dataset, showing our method could find 250 comparable traffic scenes for a handpicked car-following scenario with three other vehicles surrounding the ego vehicle. 
    \item With the extracted scenes, the variability in human responses be investigated, independent of the executed maneuver, and without the need for costly and time-consuming driving-simulator experiments.
    \item Our case study illustrates how the trajectories evolving from similar initial conditions (of 250 comparable traffic scenes) can be analyzed to show variability in operational \underline{and} tactical driver behavior. 
\end{itemize}

\bibliographystyle{ieeetr}
\bibliography{My_Collection}

\end{document}